\documentclass[11pt,a4paper]{article}
\usepackage[hyperref]{acl2018}

\usepackage{times}
\usepackage{latexsym}
\usepackage{color}
\usepackage{subcaption}
\usepackage{url}
\usepackage{graphicx}
\usepackage{mathtools} 
\usepackage[]{algorithm2e}
\usepackage{booktabs}

\aclfinalcopy 
\def\aclpaperid{457} 

\newcommand{\forwardfirst}{\textit{F$_{1st}(w)$}}
\newcommand{\backwardfirst}{\textit{B$_{1st}(w)$}}
\newcommand{\forwardlast}{\textit{F$_{last}(w)$}}
\newcommand{\backwardlast}{\textit{B$_{last}(w)$}}

\newcommand{\forwardfirstnow}{\textit{F$_{1st}$}}
\newcommand{\backwardfirstnow}{\textit{B$_{1st}$}}
\newcommand{\forwardlastnow}{\textit{F$_{last}$}}
\newcommand{\backwardlastnow}{\textit{B$_{last}$}}

\title{Morphosyntactic Tagging with a Meta-BiLSTM Model\\ over Context Sensitive Token Encodings\\
}

\author{Bernd Bohnet, Ryan McDonald, Gon\c calo Sim\~oes, Daniel Andor,  Emily Pitler, Joshua Maynez \\
  Google Inc. \\
  {\{\tt bohnetbd,ryanmcd,gsimoes,andor,epitler,joshuahm\}@google.com} }

\date{}

\begin{document}
\maketitle
\begin{abstract}

The rise of neural networks, and particularly recurrent neural networks, has produced significant advances in part-of-speech tagging accuracy \cite{zeman-EtAl:2017:K17-3}.
One characteristic common among these models is the presence of rich initial word encodings.
These encodings typically are composed of a recurrent character-based representation with learned and pre-trained word embeddings.
However, these encodings do not consider a context wider than a single word and it is only through subsequent recurrent layers that word or sub-word information interacts.
In this paper, we investigate models that use recurrent neural networks with sentence-level context for initial character and word-based representations. In particular we show that optimal results are obtained by integrating these context sensitive representations through synchronized training with a meta-model that learns to combine their states.
We present results on part-of-speech and morphological tagging with state-of-the-art performance on a number of languages. 

\end{abstract}

\section{Introduction}

Morphosyntactic tagging accuracy has seen dramatic improvements through the adoption of recurrent neural networks---specifically BiLSTMs \cite{schuster1997bidirectional,graves2005framewise} to create sentence-level context sensitive encodings of words. A successful recipe is to first create an initial context insensitive word representation, which usually has three main parts: 1) A dynamically trained word embedding; 2) a fixed pre-trained word-embedding, induced from a large corpus; and 3) a sub-word character model, which itself is usually the final state of a recurrent model that ingests one character at a time. Such word/sub-word models originated with \newcite{DBLP:journals/corr/PlankSG16}. Recently, \newcite{DBLP:conf/conll/DozatQM17} used precisely such a context insensitive word representation as input to a BiLSTM in order to obtain context sensitive word encodings used to predict part-of-speech tags. The Dozat et al.\ model had the highest accuracy of all participating systems in the CoNLL 2017 shared task \cite{zeman-EtAl:2017:K17-3}.

In such a model, sub-word character-based representations only interact indirectly via subsequent recurrent layers. For example, consider the sentence \emph{I had shingles, which is a painful disease}. 
Context insensitive character and word representations may have learned that for unknown or infrequent words like `shingles', `s' and more so `es' is a common way to end a plural noun. It is up to the subsequent BiLSTM layer to override this once it sees the singular verb is to the right. Note that this differs from traditional linear models where word and sub-word representations are directly concatenated with similar features in the surrounding context \cite{gimenez2004}.

In this paper we aim to investigate to what extent having initial sub-word and word context insensitive representations affects performance. We propose a novel model where we learn context sensitive initial character and word representations through two separate sentence-level recurrent models. These are then combined via a meta-BiLSTM model that builds a unified representation of each word that is then used for syntactic tagging. Critically, while each of these three models---character, word and meta---are trained synchronously, they are ultimately separate models using different network configurations, training hyperparameters and loss functions. Empirically, we found this optimal as it allowed control over the fact that each representation has a different learning capacity.

We tested the system on the 2017 CoNLL shared task data sets and gain improvements compared to the top performing systems for the majority of languages for part-of-speech and morphological tagging. As we will see, a pattern emerged where gains were largest for morphologically rich languages, especially those in the Slavic family group.
We also applied the approach to the benchmark English PTB data, where our model achieved 97.9 using the standard train/dev/test split, which constitutes a relative reduction in error of 12\% over the previous best system.

\section{Related Work}

While sub-word representations are often attributed to the advent of deep learning in NLP, it was, in fact, commonplace for linear featurized machine learning methods to incorporate such representations. While the literature is too large to enumerate, 
\newcite{gimenez2004} is a good example of an accurate linear model that uses both word and sub-word features. Specifically, like most systems of the time, they use n-gram affix features, which were made context sensitive via manually constructed conjunctions with features from other words in a fixed window.

\newcite{collobert2008unified} was perhaps the first modern neural network for tagging. While this first study used only word embeddings, a subsequent model extended the representation to include suffix embeddings \cite{collobert2011natural}.

The seminal dependency parsing paper of \newcite{chen2014fast} led to a number of tagging papers that used their basic architecture of highly featurized (and embedded) feed-forward neural networks. \newcite{botha2017natural}, for example, studied this architecture in a low resource setting using word, sub-word (prefix/suffix) and induced cluster features to obtain competitive accuracy with the state-of-the-art. \newcite{zhou-zhang-huang-chen:2015:ACL}, \newcite{alberti2015improved} and \newcite{andor2016globally} extended the work of Chen et al.\  to a structured prediction setting, the later two use again a mix of word and sub-word features.

The idea of using a recurrent layer over characters to induce a complementary view of a word has occurred in numerous papers. Perhaps the earliest is \newcite{santos2014learning} who compare character-based LSTM encodings to traditional word-based embeddings. \newcite{ling2015finding} take this a step further and combine the word embeddings with a recurrent character encoding of the word---instead of just relying on one or the other. \newcite{alberti17} use a sentence-level character LSTM encoding for parsing. \newcite{Petters18} show that contextual embeddings using character convolutions improve accuracy for number of NLP tasks.
\newcite{DBLP:journals/corr/PlankSG16} is probably the jumping-off point for most current architectures for tagging models with recurrent neural networks. Specifically, they used a combined word embedding and recurrent character encoding as the initial input to a BiLSTM that generated context sensitive word encodings. Though, like most previous studies, these initial encodings were context insensitive and relied on subsequent layers to encode sentence-level interactions.

Finally, \newcite{DBLP:conf/conll/DozatQM17} showed that sub-word/word combination representations lead to state-of-the-art morphosyntactic tagging accuracy across a number of languages in the CoNLL 2017 shared task \cite{zeman-EtAl:2017:K17-3}. Their word representation consisted of three parts: 1) A dynamically trained word embedding; 2) a fixed pre-trained word embedding; 3) a character LSTM encoding that summed the final state of the recurrent model with vector constructed using an attention mechanism over all character states. Again, the initial representations are all context insensitive. As this model is currently the state-of-the-art in morphosyntactic tagging, it will serve as a baseline during our discussion and experiments.

\section{Models}
\label{sec:models}

In this section, we introduce models that we investigate and experiment with in \S\ref{sec:experiments}.

\subsection{Sentence-based Character Model}

\begin{figure*}
\centering
\begin{subfigure}[t]{0.47\textwidth}
    \includegraphics[width=\textwidth]{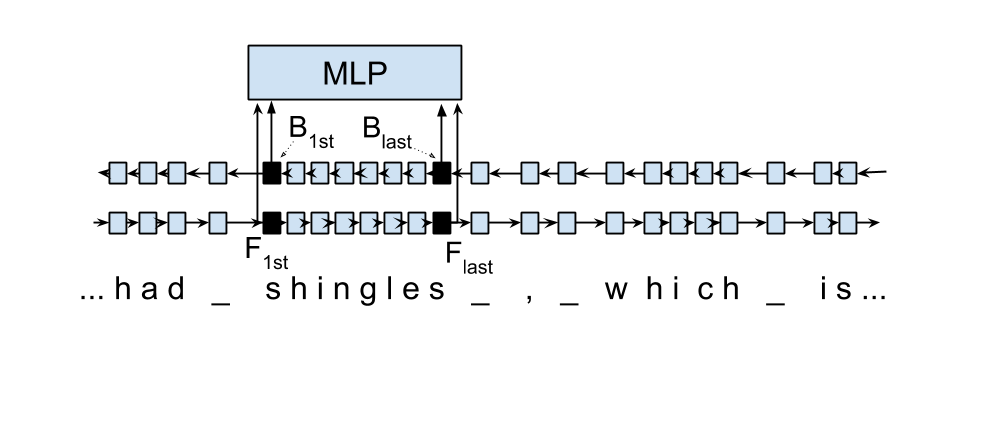}
    \vspace{-0.5in}
    \caption{Sentence-based Character Model.  The representation for the token \emph{shingles} is the concatenation of the four shaded boxes.  Note the surrounding sentence context affects the representation.}
    \label{fig:charmodel}
\end{subfigure}
\hspace{.3cm}
\begin{subfigure}[t]{0.47\textwidth}
\includegraphics[width=\textwidth]{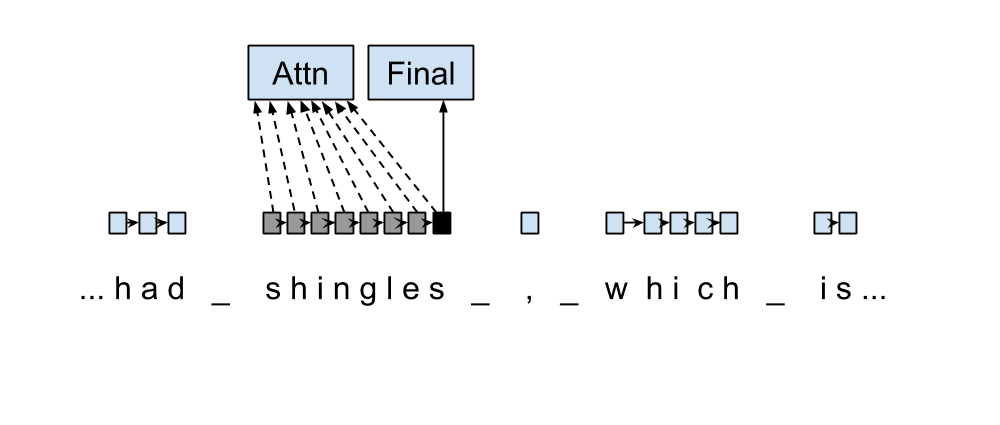}
\vspace{-0.5in}
\caption{Token-based Character Model\footnote{This is specifically the model of \newcite{DBLP:conf/conll/DozatQM17}.}.  The token is represented by the concatenation of attention over the lightly shaded boxes with the final cell (dark shaded box).  The rest of the sentence has no impact on the representation.}
\label{fig:tokencharmodel}
\end{subfigure}
\caption{Token representations
are sensitive to the context
in the sentence-based character model (\S\ref{sec:chars-model})
and are context-independent
in the token-based character model (\S\ref{sec:word-based}).}
\label{fig:comparingcharmodels}
\end{figure*}

The feature that distinguishes our model most from previous work is that we apply a bidirectional recurrent layer (LSTM) on all characters of a sentence to induce fully context sensitive initial word encodings. 
That is, we do not restrict the context of this layer to the words themselves (as in Figure \ref{fig:tokencharmodel}). 
Figure \ref{fig:charmodel} shows the sentence-based character
model applied to an example token in context.


\label{sec:chars-model}

The character model uses, as input, sentences split into UTF8 characters. We include the spaces between the tokens\footnote{As input, we assume the sentence has been tokenized/segmented.} in the input and map each character to a dynamically learned embedding.

Next, a forward LSTM reads the characters from left to right and a backward LSTM reads sentences from right to left, in standard BiLSTM fashion.

More formally, for an $n$-character sentence, we apply for each character embedding ($e_1^{char}, ..., e_n^{char}$) a BiLSTM:     
\begin{eqnarray*}
\nonumber
f_{c,i}^0, b_{c,i}^0 &=&\textrm{BiLSTM}(r_0,(e_1^{char}, ..., e_n^{char}))_i 
\end{eqnarray*}

As is also typical, we can have multiple such layers ($l$) that feed into each other through the concatenation of previous layer encodings.

The last layer $l$ has both forward $(f_{c,1}^l, ..., f_{c,n}^l)$ and backward $(b_{c,1}^l, ..., b_{c,n}^l)$ output vectors for each character. 

To create word encodings, we need to
combine a relevant subset of these context sensitive character encodings.
These word encodings can then be used in a model that assigns morphosyntactic tags to each word directly or via subsequent layers.
To accomplish this, the model concatenates up to four character output vectors: the \{\emph{forward}, \emph{backward}\} output of the \{\emph{first}, \emph{last}\} character in the token ($\forwardfirst$, $\forwardlast$, $\backwardfirst$, $\backwardlast$).  In Figure \ref{fig:charmodel}, the four
shaded boxes indicate these four outputs for the example token.

Thus, the proposed model concatenates all four of these and passes it as input to an multilayer perceptron (MLP):
\begin{eqnarray}
g_i & = & \textrm{concat}(\forwardfirst, \forwardlast, \nonumber  \\
& & \quad \backwardfirst, \backwardlast)  \label{eq:gi} \\
m_i^{chars} &=&\textrm{MLP}(g_i) \label{equation:mlp-char}\nonumber  
\end{eqnarray}
A tag can then be predicted with a linear classifier 
that takes as input the output of the MLP $m_i^{chars}$, applies a softmax function and chooses for each word the tag with highest probability. 
Table \ref{table:char-models} investigates
the empirical impact of alternative
definitions of $g_i$ that concatenate only subsets of $\{\forwardfirst, \forwardlast, \backwardfirst, \backwardlast\}$.

\subsection{Word-based Character Model}
\label{sec:word-based}

To investigate whether a sentence sensitive character model is better than a character model where the context is restricted to the characters of a word, we reimplemented the word-based character model of \newcite{DBLP:conf/conll/DozatQM17} as shown in  Figure~\ref{fig:charmodel}. This model uses the final state of a unidirectional LSTM over the characters of the word, combined with the attention mechanism of \newcite{CaoRei16} over all characters. We refer the reader to those works for more details. Critically, however, all the information fed to this representation comes from the word itself, and not a wider sentence-level context.

\subsection{Sentence-based Word Model}
\label{sec:word-model}

We used a similar setup for our context sensitive word encodings as the character encodings. There are a few differences. Obviously, the inputs are the words of the sentence. For each of the words, we use pretrained word embeddings ($p_1^{word}, ..., p_n^{word}$) summed with a dynamically learned word embedding for each word in the corpus ($e_1^{word}, ..., e_n^{word}$):
\[in_i^{word} = e_i^{word}~+~p_i^{word}\]
The summed embeddings $in_i$ are passed as input to one or more BiLSTM layers whose output $f_{w,i}^l, b_{w,i}^l$ is concatenated and used as the final encoding, which is then passed to an MLP 

\begin{eqnarray}
\nonumber
\left.\begin{aligned}
o_i^{word} &=\textrm{concat}(f_{w,i}^l, b_{w,i}^l) \\ \nonumber
m_i^{word} &=\textrm{MLP}(o_i^{word}) \label{equation:mlp-word}\\
\end{aligned}\right.
\end{eqnarray}
It should be noted, that the output of this BiLSTM is essentially the Dozat et al.\ model before tag prediction, with the exception that the word-based character encodings are excluded.


\begin{figure*}[t!]
\begin{subfigure}[t]{0.5\textwidth}
\centering\includegraphics[clip, trim=8cm 7.5cm 9cm 2.5cm, width=0.85\textwidth]{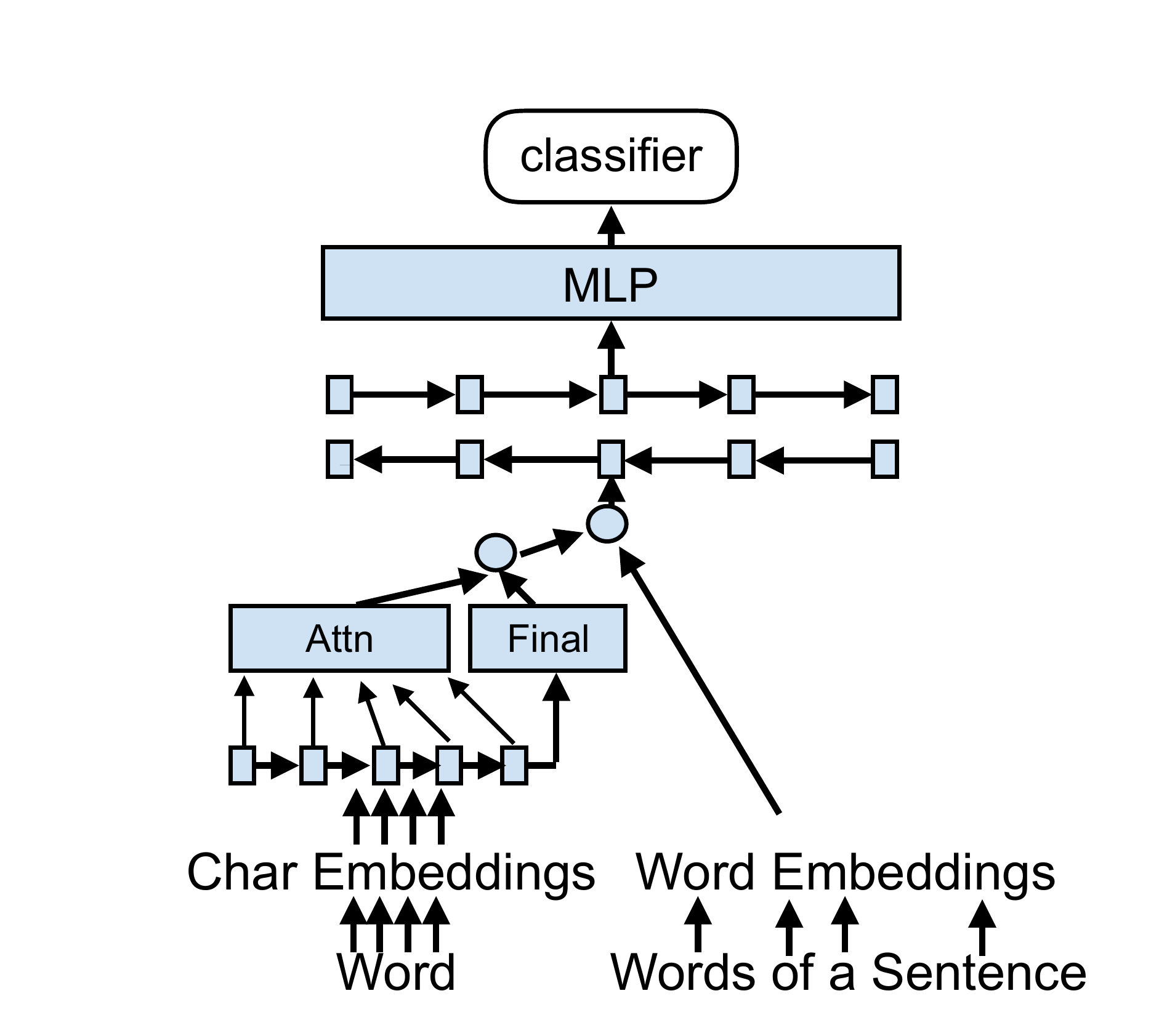}
\caption{The overall architecture of \newcite{DBLP:conf/conll/DozatQM17}}
\label{figure:architecture-dm}
\end{subfigure}
\begin{subfigure}[t]{0.5\textwidth}
\centering\includegraphics[clip, trim=8cm 7cm 8cm 2.5cm, width=0.85\textwidth]{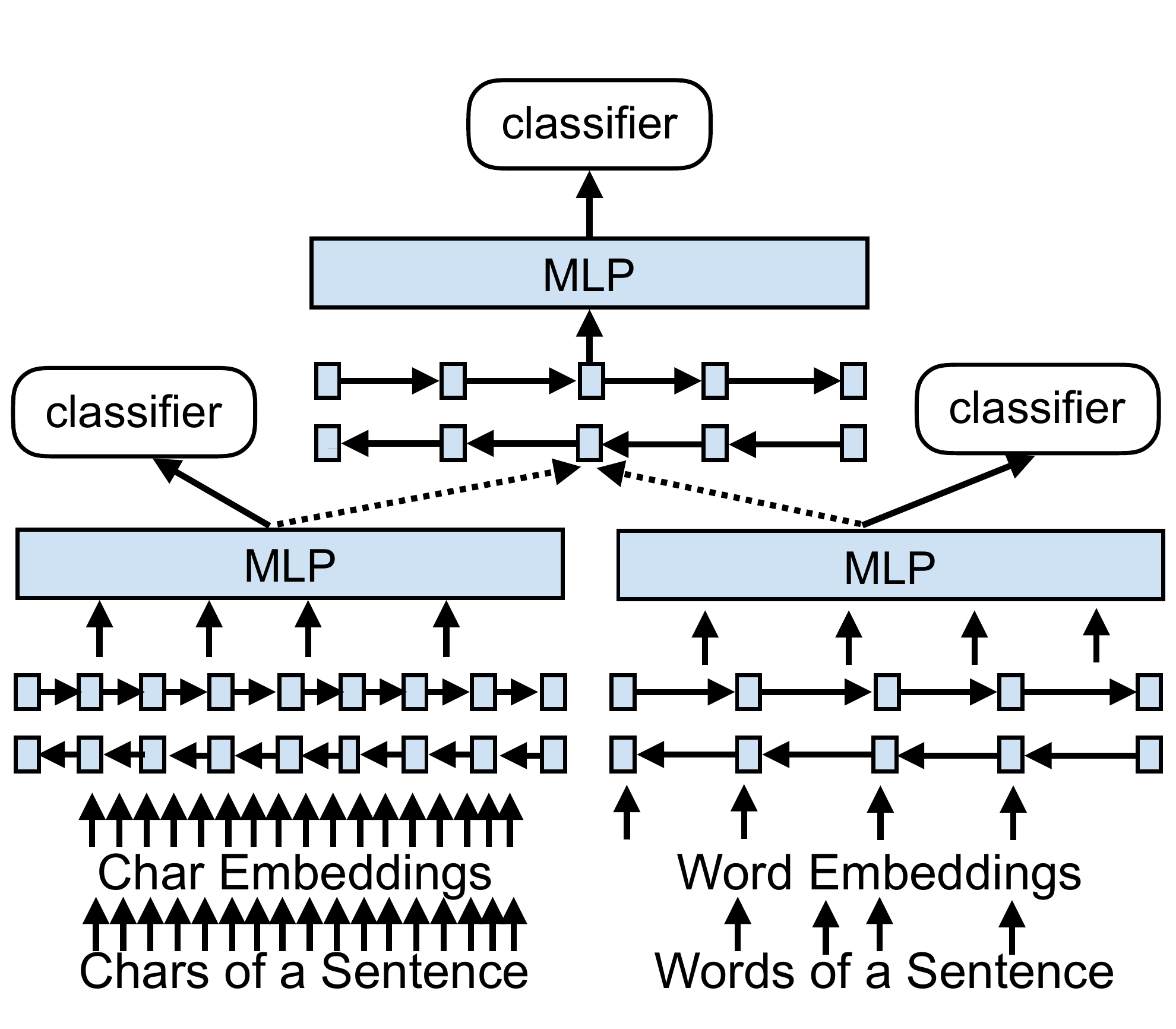}
\caption{The overall architecture of the system. The data flows along the arrows. The optimizers minimizes the loss of the classifiers independently and backpropagates along the bold arrows.}
\label{figure:architecture}
\end{subfigure}
\caption{Tagging architectures. (a) \newcite{DBLP:conf/conll/DozatQM17}; (b) Meta-BiLSTM architecture of this work.}
\end{figure*}

\subsection{Meta-BiLSTM: Model Combination}

Given initial word encodings, both character and word-based, a common strategy is to pass these through a sentence-level BiLSTM to create context sensitive encodings, e.g., this is precisely what \newcite{DBLP:journals/corr/PlankSG16} and \newcite{DBLP:conf/conll/DozatQM17} do. However, we found that if we trained each of the character-based and word-based encodings with their own loss, and combined them using an additional meta-BiLSTM model, we obtained optimal performance. In the meta-BiLSTM model, we concatenate the output, for each word, of its context sensitive character and word-based encodings, and put this through another BiLSTM to create an additional combined context sensitive encoding. This is followed by a final MLP whose output is passed to a linear layer for tag prediction. 
\begin{eqnarray}
\nonumber
\left.\begin{aligned}
cw_i &=\textrm{concat}(m_i^{char}, m_i^{word})  \\ \nonumber
f_{m,i}^l, b_{m,i}^l &=\textrm{BiLSTM}(r_0,(cw_0, ..., cw_n))_i \\ \nonumber
m_i^{comb} &=\textrm{MLP}(\textrm{concat}(f_{m,i}^l, b_{m,i}^l))
\end{aligned}\right.
\end{eqnarray}
With this setup, each of the models can be optimized independently which we describe in more detail in \S\ref{sec:schema}. Figure~\ref{figure:architecture} depicts the architecture of the combined system and contrasts it with that of the Dozat et al.\ model (Figure~\ref{figure:architecture-dm}).

\subsection{Training Schema}
\label{sec:schema}

As mentioned in the previous section, the character and word-based encoding models have their own tagging loss functions, which are trained independently and joined via the meta-BiLSTM. I.e., the loss of each model is minimized independently by separate optimizers with their own hyperparameters. 
Thus, it is in some sense a multi-task learning model and we must define a schedule in which individual models are updated. We opted for a simple synchronous schedule outline in Algorithm~\ref{alg:training}. Here, during each epoch, we update each of the models in sequence---character, word and meta---using the entire training data.

\begin{algorithm}[ht!]
\small
\SetNoFillComment
\newcommand{\var}[1]{\textit{#1}}
\newcommand{\kw}[1]{\textbf{#1}}
\newcommand\mycommfont[1]{\footnotesize\ttfamily{#1}}
\SetCommentSty{mycommfont}
 \kw{Data:}~{train-corpus, dev-corpus}\\
 \tcc{The following models are defined in \S\ref{sec:models}.}
 \textbf{Input:} char-model, word-model, meta-model\\
 \tcc{Model optimizers}
 \textbf{Input:} char-opt, word-opt, meta-opt \\
 \tcc{Results are parameter sets for each model.}
 \kw{Result:}~{best-char, best-word, best-meta}\\
 \tcc{Initialize parameter sets (cf. Table~\ref{table:hyper-params})}
 Initialize($pa_c, pa_w, pa_m$)\\
 \tcc{Iteration on over training corpus.} 
 \For{epoch = 1 \kw{to} MAX}  { 
   \tcc{Update character model.}
   char-logits, char-preds =  char-model(train-corpus, $pa_c$)
   $pa_c$ = char-opt.update(char-preds, train-data)
   \tcc{Update word model.} 
   word-logits, word-preds =  word-model(train-corpus, $pa_w$)
   $pa_w$ = word-opt.update(char-preds, train-data)
   \tcc{Update Meta-BiLSTM}
   meta-preds = meta-model(train-corpus, $pa_c, pa_w$, $pa_m$)
   $pa_m$ = meta-opt.update(train-corpus, meta-preds)
  \tcc{Evaluate model due to dev set accuracy.}
  F1 = DevEval($par_c$, $par_w$, $par_m$)
  \tcc{Keep the best model.}
  \If{F1 $>$ best-F1 } {
    \var{best-char} = $pa_c$; \var{best-word} = $pa_w$ 
    \var{best-meta} = $pa_m$; \var{best-F1} = \var{F1}
  }
}
 
 \caption{Training procedure for learning initial character and word-based context sensitive encodings synchronously with meta-BiLSTM.}
 \label{alg:training}
\end{algorithm}

In terms of model selection, after each epoch, the algorithm evaluates the tagging accuracy of the development set and keeps the parameters of the best model. Accuracy is measured using the meta-BiLSTM tagging layer, which requires a forward pass through all three models. Though we use all three losses to update the models, only the meta-BiLSTM layer is used for model selection and test-time prediction.

While each of the three models---character, word and meta---are trained with their own loss functions, it should be emphasized that training is synchronous in the sense that the meta-BiLSTM model is trained in tandem with the two encoding models, and not after those models have converged. Since accuracy from the meta-BiLSTM model on the development set determines the best parameters, training is not completely independent. We found this to improve accuracy overall. Crucially, when we allowed the meta-BiLSTM to back-propagate through the whole network, performance degraded regardless of whether one or multiple loss functions were used.

Each language could in theory use separate hyperparameters, optimized for highest accuracy. However, for our main experiments we use identical settings for each
language 
which worked well for large corpora and simplified things. We provide an overview of the selected hyperparameters in \S\ref{sec:hyperparameters}. We explored more settings for selected individual languages with a grid search and ablation experiments and present the results in \S\ref{sec:discussion}.

\section{Experiments and Results}
\label{sec:experiments}

In this section, we present the experimental setup and the selected hyperparameter for the main experiments where we use the CoNLL Shared Task 2017 treebanks and compare with the best systems of the shared task.  

\subsection{Experimental Setup}
\label{sec:hyperparameters}

For our main results, we selected one network configuration and set of the hyperparameters. These settings are not optimal for all languages. However, since hyperparameter exploration is computationally demanding due to the number of languages we optimized these hyperparameters on initial development data experiments over a few languages. Table~\ref{table:hyper-params} shows an overview of the architecture, hyperparameters and the initialization settings of the network.
The word embeddings are initialized with zero values and the pre-trained embeddings are not updated during training. The dropout used on the embeddings is achieved by a single dropout mask and we use dropout on the input and the states of the LSTM.

As is standard, model selection was done measuring development accuracy/F1 score after each epoch and taking the model with maximum value on the development set.

\begin{table}[t!]
\begin{center}
\small
\begin{tabular}{|l|l|l|}
\hline 
\multicolumn{3}{|c|}{\bf Architecture } \\ \hline
\bf Model          &\bf Parameter        & \bf Value \\ \hline
Chr, Wrd           & BiLSTM layers      & 3 \\
Mt                 & BiLSTM layers      & 1 \\
Chr, Wrd, Mt       & BiLSTM size        & 400 \\
Chr, Wrd, Mt       & Dropout LSTMs      & 0.33 \\
Chr, Wrd, Mt       & Dropout MLP        & 0.33 \\
Wrd                & Dropout embeddings & 0.33 \\ 
Chr                & Dropout embeddings & 0.05 \\ 
Chr, Wrd, Mt       & Nonlinear act. (MLP) & ELU \\ \hline
\hline 
\multicolumn{3}{|c|}{\bf Initialization } \\ \hline
\bf Model          &\bf Parameter &\bf Value \\ 
Wrd                & embeddings        & Zero \\
Chr                & embeddings        & Gaussian \\
Chr, Wrd, Mt       & MLP               & Gaussian\\ \hline
\hline 
\multicolumn{3}{|c|}{\bf Training } \\ \hline
\bf Model          &\bf Parameter        & \bf Value \\ \hline
Chr, Wrd, Mt       & Optimizer          & Adam\\
Chr, Wrd, Mt       & Loss               & Cross entropy\\
Chr, Wrd, Mt       & Learning rate      & 0.002\\
Chr, Wrd, Mt       & Decay              & 0.999994\\
Chr, Wrd, Mt       & Adam epsilon       & 1e-08\\
Chr, Wrd, Mt       & beta1              & 0.9\\
Chr, Wrd, Mt       & beta2              & 0.999\\\hline
\end{tabular}
\end{center}
\caption{\label{table:hyper-params} Selected hyperparameters and initialization of parameters for our models. \emph{Chr}, \emph{Wrd}, and \emph{Mt} are used to indicate the character, word, and meta models respectively.
The Gaussian distribution is used with a mean of 0 and variance of 1 to generate the random values. 
}
\end{table}

\begin{table}[]
\begin{center}
\small
\renewcommand{\arraystretch}{.95}
\begin{tabular}{|l|r||r|r|r|r|}
\hline 
            &    CONLL    & \bf{DQM}   & \bf ours   &\bf RRIE \\
\bf lang.   &    Winner   &            &            &         \\ \hline
cs\_cac     &      95.16  &     95.16  & \bf 96.91  &     36.2\\
cs          &      95.86  &     95.86  & \bf 97.28  &     35.5\\
fi          &      97.37  &     97.37  & \bf 97.81  &     16.7\\
sl          &      94.74  &     94.74  & \bf 95.54  &     15.2\\
la\_ittb    &      94.79  &     94.79  & \bf 95.56  &     14.8\\
grc         &      84.47  &     84.47  & \bf 86.51  &     13.1\\
bg          &      96.71  &     96.71  & \bf 97.05  &     10.3\\
ca          &      98.58  &     98.58  & \bf 98.72  &      9.9\\
grc\_proiel &      97.51  &     97.51  & \bf 97.72  &      8.4\\
pt          &      83.04  &     83.04  & \bf 84.39  &      8.0\\
cu          &      96.20  &     96.20  & \bf 96.49  &      7.6\\
it          &      97.93  &     97.93  & \bf 98.08  &      7.2\\
fa          &      97.12  &     97.12  & \bf 97.32  &      6.9\\
ru          &      96.73  &     96.73  & \bf 96.95  &      6.7\\
sv          &      96.40  &     96.40  & \bf 96.64  &      6.7\\
ko          &      93.02  &     93.02  & \bf 93.45  &      6.2\\
sk          &      85.00  &     85.00  & \bf 85.88  &      5.9\\
nl          &      90.61  &     90.61  & \bf 91.10  &      5.4\\
fi\_ftb     &      95.31  &     95.31  & \bf 95.56  &      5.3\\
de          &      97.29  &     97.29  & \bf 97.39  &      4.7\\
tr          &      93.11  &     93.11  & \bf 93.43  &      4.6\\
hi          &      97.01  &     97.01  & \bf 97.13  &      4.0\\
es\_ancora  &      98.73  &     98.73  & \bf 98.78  &      3.9\\
ro          &      96.98  &     96.98  & \bf 97.08  &      3.6\\
la\_proiel  &      96.93  &     96.93  & \bf 97.00  &      2.3\\
pl          &      91.97  &     91.97  & \bf 92.12  &      1.9\\
ar          &      87.66  &     87.66  & \bf 87.82  &      1.3\\
gl          &      97.50  &     97.50  & \bf 97.53  &      1.2\\
sv\_lines   &      94.84  &     94.84  & \bf 94.90  &      1.2\\
cs\_clt     &      89.98  &     89.98  & \bf 90.09  &      1.1\\
lv          &      80.05  &     80.05  & \bf 80.20  &      0.8 \\
zh          & \bf  88.40  &     85.07  &     85.10  &      0.2\\
da          & \bf 100.00  &     99.96  &     99.96  &      0.0\\
es          & \bf  99.81  &     99.69  &     99.69  &      0.0\\
eu          & \bf  99.98  &     99.96  &     99.96  &      0.0\\
fr\_sequoia & \bf  99.49  &     99.06  &     99.06  &      0.0\\
fr          & \bf  99.50  &     98.87  &     98.87  &      0.0\\
hr          & \bf  99.93  & \bf 99.93  & \bf 99.93  &      0.0\\
hu          & \bf  99.85  &     99.82  &     99.82  &      0.0\\
id          & \bf 100.00  &     99.99  &     99.99  &      0.0\\
ja          & \bf  98.59  &     89.68  &     89.68  &      0.0\\
nl\_lassy   & \bf  99.99  &     99.93  &     99.93  &      0.0\\
no\_bok.    & \bf  99.88  &     99.75  &     99.75  &      0.0\\
no\_nyn.    & \bf  99.93  &     99.85  &     99.85  &      0.0\\
ru\_syn.    & \bf  99.58  &     99.57  &     99.57  &      0.0\\
en\_lines   & \bf  95.41  & \bf 95.41  &     95.39  &     -0.4\\
ur          & \bf  92.30  & \bf 92.30  &     92.21  &     -1.2\\
he          & \bf  83.24  &     82.45  &     82.16  &     -1.7\\
vi          & \bf  75.42  &     73.56  &     73.12  &     -1.7\\
gl\_treegal & \bf  91.65  & \bf 91.65  &     91.40  &     -3.0\\
en          & \bf  94.82  & \bf 94.82  &     94.66  &     -3.1\\
en\_partut  & \bf  95.08  & \bf 95.08  &     94.81  &     -5.5\\
pt\_br      & \bf  98.22  & \bf 98.22  &     98.11  &     -6.2\\
et          & \bf  95.05  & \bf 95.05  &     94.72  &     -6.7\\
el          & \bf  97.76  & \bf 97.76  &     97.53  &    -10.3\\\hline
macro-avg   &      93.18  &     93.11  & \bf 93.40  &      -  \\\hline
\end{tabular}
\end{center}
    \caption{\label{table:state-of-the-art-xpos} Results for XPOS tags. The first column shows the language acronym, the column named \textbf{DQM} shows the results of \newcite{DBLP:conf/conll/DozatQM17}. Our system outperforms \newcite{DBLP:conf/conll/DozatQM17} on 32 out of 54 treebanks and Dozat et al. outperforms our model on 10 of 54 treebanks, with 13 ties. \textbf{RRIE} is the relative reduction in error. We excluded ties in the calculation of macro-avg since these treebanks do not contain meaningful xpos tags.} 
\end{table}

\subsection{Data Sets}

For the experiments, we use the data sets as provided by the CoNLL Shared Task 2017 \cite{zeman-EtAl:2017:K17-3}. For training, we use the training sets which were denoted as big treebanks \footnote{In the CONLL 2017 Shared Task, a big treebank is one that contains a development set. In total, there are 55 out of the 64 UD treebanks which are considered big treebanks.}.

We followed the same methodology used in the CoNLL Shared Task. We use the training treebank for training only and the development sets for hyperparameter tuning and early stopping. To keep our results comparable with the Shared Task, we use the provided precomputed word embeddings. We excluded Gothic from our experiments as the available downloadable content did not include embeddings for this language.

As input to our system---for both part-of-speech tagging and morphological tagging---we use the output of the UDPipe-base baseline system \cite{11234/1-1990} which provides segmentation. The segmentation differs from the gold segmentation and impacts accuracy negatively for a number of languages. Most of the top performing systems for part-of-speech tagging used as input UDPipe to obtain the segmentation for the input data. For morphology, the top system for most languages (IMS) used its own segmentation \cite{bjorkelund-EtAl:2017:K17-3}. For the evaluation, we used the official evaluation script \cite{zeman-EtAl:2017:K17-3}.

\subsection{Part-of-Speech Tagging Results}

In this section, we present the results of the application of our model to part-of-speech tagging. In our first experiment, we used our model in the setting of the CoNLL 2017 Shared Task to annotate words with XPOS\footnote{These are the language specific fine-grained part-of-speech tags from the Universal Dependency Treebanks.} tags~\cite{zeman-EtAl:2017:K17-3}. We compare our results against the top systems of the CoNLL 2017 Shared Task. Table~\ref{table:state-of-the-art-xpos} contains the results of this task for the large treebanks.

Because \newcite{DBLP:conf/conll/DozatQM17} won the challenge for the majority of the languages, we first compare our results with the performance of their system. Our model outperforms~\newcite{DBLP:conf/conll/DozatQM17} in 32 of the 54 treebanks with 13 ties. These ties correspond mostly to languages where XPOS tagging anyhow obtains accuracies above 99\%. Our model tends to produce better results, especially for morphologically rich languages (e.g.~Slavic languages), whereas \newcite{DBLP:conf/conll/DozatQM17} showed higher performance in 10 languages in particular English, Greek, Brazilian Portuguese and Estonian.

\subsection{Part-of-Speech Tagging on WSJ}

We also performed experiments on the Penn Treebank  with the usual split in train, development and test set. Table~\ref{table:state-of-the-art-wsj} shows the results of our model in comparison to the results reported in state-of-the-art literature. Our model significantly outperforms these systems, with an absolute difference of 0.32\% in accuracy, which corresponds to a RRIE of 12\%.

\begin{table}[t!]
\begin{center}
\begin{tabular}{|l|l|l|l|}
\hline 
\renewcommand{\arraystretch}{0.8}
\bf System  & \bf Accuracy     \\ \hline
\newcite{Sogaard:2011:SCN}  & 97.50 \\
\newcite{DBLP:journals/corr/HuangXY15} & 97.55 \\
\newcite{choi:16a} & 97.64         \\
\newcite{andor2016globally}. & 97.44  \\
\newcite{DBLP:conf/conll/DozatQM17} & 97.41 \\
ours      & \bf 97.96 \\ \hline
\end{tabular}
\end{center}
\vspace{-0.1in}
\caption{\label{table:state-of-the-art-wsj} Results on WSJ test set.  }
\end{table}

\subsection{Morphological Tagging Results}

In addition to the XPOS tagging experiments, we performed experiments with morphological tagging. This annotation was part of the CONLL 2017 Shared Task and the objective was to predict a bundle of morphological features for each token in the text. 
Our model treats the morphological bundle as one tag making the problem equivalent to a sequential tagging problem.
Table~\ref{table:morphology-results} shows the results.

\begin{table}[]
\begin{center}
\small
\renewcommand{\arraystretch}{.92}
\begin{tabular}{|l|c||c|c|r|}
\hline 
             &   CONLL    &\bf{DQM}      & \bf ours  &\bf RRIE\\
\bf lang.    &   Winner   & Reimpl.      &           &        \\ \hline
cs\_cac      &     90.72  &     94.66    & \bf 96.41 &  27.9\\
ru\_syn.     &     94.55  &     96.70    & \bf 97.53 &  23.1\\
cs           &     93.14  &     96.32    & \bf 97.14 &  22.3\\
la\_ittb     &     94.28  &     96.45    & \bf 97.12 &  18.9\\
sl           &     90.08  &     95.26    & \bf 96.03 &  16.2\\
ca           &     97.23  &     97.85    & \bf 98.13 &  13.0\\
fi\_ftb      &     93.43  &     95.96    & \bf 96.42 &  11.4\\
no\_bok.     &     95.56  &     96.95    &\bf  97.26 &  10.2\\
grc\_proiel  &     90.24  &     91.35    & \bf 92.22 &  10.1\\
fr\_sequoia  &     96.10  &     96.62    & \bf 97.62 &  10.1\\
la\_proiel   &     89.22  &     91.52    & \bf 92.35 &   9.8\\
es\_ancora   &     97.72  &     98.15    & \bf 98.32 &   9.7\\
da           &     94.83  &     96.62    & \bf 96.94 &   9.5\\
fi           &     92.43  &     94.29    & \bf 94.83 &   9.5\\
sv           &     95.15  &     96.52    & \bf 96.84 &   9.2\\
pt           &     94.62  &     95.89    & \bf 96.27 &   9.2\\
grc          &     88.00  &     90.39    & \bf 91.13 &   9.0\\
no\_nyn.     &     95.25  &     96.79    &\bf  97.08 &   9.0\\
de           &     83.11  &     89.78    & \bf 90.70 &   9.0\\
ru           &     87.27  &     91.99    & \bf 92.69 &   8.7\\
hi           &     91.03  &     90.72    &\bf  91.78 &   8.1\\
cu           &     88.90  &     88.93    & \bf 89.82 &   8.0\\
fa           &     96.34  &     97.23    & \bf 97.45 &   7.9\\
tr           &     87.03  &     89.39    & \bf 90.21 &   7.7\\
en\_partut   &     92.69  &     93.93    & \bf 94.40 &   7.7\\
sk           &     81.23  &     87.54    & \bf 88.48 &   7.5\\  
eu           &     89.57  &     92.48    & \bf 93.04 &   7.4\\
pt\_br       &    99.73  &     99.73    &  \bf 99.75 &   7.4\\
es           &     96.34  &     96.42    & \bf 96.68 &   7.3\\
ko           &     99.41  &   99.44    & \bf     99.48 &   7.1\\
ar           &     87.15  &     85.45    & \bf 88.29 &   6.7\\
it           &     97.37  &  97.72    &  \bf   97.86 &   6.1\\
nl\_lassy    &     97.55  &     98.04    & \bf 98.15 &   5.2\\
nl           &     90.04  &     92.06    & \bf 92.47 &   5.2\\
pl           &     86.53  &     91.71    &\bf  92.14 &   5.2\\
ur           &     81.03  &     83.16    & \bf 84.02 &   5.1\\
bg           &     96.47  &     97.71    & \bf 97.82 &   4.8\\
hr           &     85.82  &     90.64    & \bf 91.50 &   3.8\\
he           & \bf 85.06  &     79.34    &     79.76 &   2.0\\
et           &     84.62  &     88.18    & \bf 88.25 &   0.6\\
zh           & \bf 92.90  &     87.67    &     87.74 &   0.6\\
vi           & \bf 86.92  &     82.23    &     82.30 &   0.4\\
ja           & \bf 96.84  &     89.65    &     89.66 &   0.1\\
en\_lines    &     99.96  &     99.99    &     99.99 &   0.0\\
fr           &     96.12  &     95.98    &     95.98 &   0.0\\
gl           & \bf 99.78  &     99.72    &     99.72 &   0.0\\
id           & \bf 99.55  &     99.50    &     99.50 &   0.0\\
ro           &     96.24  &     97.26    &  97.26 &   0.0\\
sv\_lines    &  99.98  &  99.98    &  99.98 &     0.0\\
cs\_cltt     &     87.88  & \bf  90.41    &  90.36 &    -0.5\\
lv           &     84.14  & \bf  87.00    &  86.92 &     -0.6\\
el           &     91.37  & \bf  94.00    &  93.92 &    -1.3\\
hu           &     72.61  & \bf 82.67    &     82.44 &  -1.3\\
en           &     94.49  & \bf 95.93    &     95.71 &  -5.4\\
\hline
macro-avg    & 91.51      &     92.89    &\bf  93.31 & - \\\hline
\end{tabular}
\end{center}
\caption{\label{table:morphology-results} Results for morphological features. The column \textbf{CoNLL Winner} shows the top system of the ST 17, the \textbf{DQM Reimpl.} shows our reimplementation of \newcite{DBLP:conf/conll/DozatQM17}, the column \textbf{ours} shows our system with a sentence-based character model; \textbf{RRIE} gives the relative reduction in error between the Reimpl. DQM and sentence-based character system. Our system outperforms the CoNLL Winner by 48 out of 54 treebanks and the reimplementation of DQM, by 43 of 54 treebanks, with 6 ties. 
}
\end{table}

Our models tend to produce significantly better results than the winners of the CoNLL 2017 Shared Task (i.e., 1.8\% absolute improvement on average, corresponding to a RRIE of 21.20\%). The only cases for which this is not true are again languages that require significant segmentation efforts (i.e., Hebrew, Chinese, Vietnamese and Japanese) or when the task was trivial.

Given the fact that~\newcite{DBLP:conf/conll/DozatQM17} obtained the best results in part-of-speech tagging by a significant margin in the CoNLL 2017 Shared Task, it would be expected that their model would also perform significantly well in morphological tagging since the tasks are very similar. Since they did not participate in this particular challenge, we decided to reimplement their system to serve as a strong baseline. 
As expected, our reimplementation of~\newcite{DBLP:conf/conll/DozatQM17} tends to significantly outperform the winners of the CONLL 2017 Shared Task. However, in general, our models still obtain better results, outperforming Dozat et al.\ on 43 of the 54 treebanks, with an absolute difference of 0.42\% on average.

\section{Ablation Study}

The model proposed in this paper of a Meta-BiLSTM with a sentence-based character model differs from prior work in multiple aspects.
In this section, we perform ablations to determine the relative impact of each modeling decision.

For the experimental setup of the ablation experiments, we report accuracy scores for the development sets. We split off 5\% of the sentences from each training corpus and we use this part for early stopping. Ablation experiments are either performed on a few selected treebanks to show individual language results or averaged across all treebanks for which tagging is non-trivial.

\paragraph{Impact of the Training Schema}
We first compare jointly training the three model components (Meta-BiLSTM, character model, word model)  to training each separately.
Table \ref{tab:ablation-opt-sep-full} shows that separately optimized models are significantly more accurate on average than jointly optimized models. 
Separate optimization leads to better accuracy for 34 out of 40 treebanks for the morphological features task and for 30 out of 39 treebanks for xpos tagging. Separate optimization out-performed joint optimization by up to 2.1 percent absolute, while joint never out-performed separate by more than 0.5\% absolute.
We hypothesize that separately training the models forces each sub-model (word and character) to be strong enough to make high accuracy predictions and in some sense serves as a regularizer in the same way that dropout does for individual neurons.

 optimization.
\begin{table}[ht!]
\begin{center}
\small
\setlength{\tabcolsep}{3.5pt}
\begin{tabular}{|l|c|c|c|c|c|c}
\hline 
 \bf Optimization   & Avg. F1 Score   & Avg. F1 Score \\ 
                    &  morphology &  xpos \\ \hline
separate  & \bf 94.57 &  \bf 94.85   \\
jointly   &     94.15  &  94.48    \\
 \hline
\end{tabular}
\end{center}
\caption{Comparison of optimization methods: Separate optimization of the word, character and meta model is more accurate on average than full back-propagation using a single loss function.The results are statistically significant with two-tailed paired t-test for xpos with p$<$0.001 and for morphology with $p<$0.0001.}
\label{tab:ablation-opt-sep-full}
\end{table}

\paragraph{Impact of the Sentence-based Character Model} 
We compared the setup with sentence-based character context  (Figure \ref{fig:charmodel}) to word-based character context (Figure \ref{fig:tokencharmodel}). We selected for these experiments a number of morphological rich languages. The results are shown in Table~\ref{tab:ablation-char-sent-vs-word}. The accuracy of the word-based character model joint with a word-based model were significantly lower than a sentence-based character model. We conclude also from these results and comparing with results of the reimplementation of DQM that early integration of the word-based character model performs much better as late integration via Meta-BiLSTM for a word-based character model.

\begin{table}[t]
\begin{center}
\small
\setlength{\tabcolsep}{3.5pt}
\begin{tabular}{|l|c|c|c|c||c|c|c|c}
\hline 
 \bf dev. set   & word char model & sentence char model\\ 
\hline
el              &  89.05 & 93.41   \\
la\_ittb        &  93.22 & 95.69  \\
ru              &  88.94 & 92.31 \\
tr              &  87.78 & 90.77 \\
 \hline
\end{tabular}
\end{center}
\caption{F1 score for selected languages on sentence vs.~word level character models for the prediction of morphology using late integration.}
\label{tab:ablation-char-sent-vs-word}
\end{table}

\paragraph{Impact of the Meta-BiLSTM Model Combination}

The proposed model trains word and character models independently while training a joint model on top. Here we investigate the part-of-speech tagging performance of the joint model compared with the word and character models on their own (using hyperparameters from in \ref{sec:hyperparameters}). 

Table~\ref{tab:ablation-charword} shows, for selected languages, the results averaged over 10 runs in order to measure standard deviation. The examples show that the combined model has significantly higher accuracy compared with either the character and word models individually. 

\begin{table}[t!]
\begin{center}
\small
\setlength{\tabcolsep}{3.5pt}
\begin{tabular}{|l|c|c|c|c||c|c|c|c}
\hline 
 \bf dev. set   & num.      & mean & mean & mean & stdev & stdev & stdev\\ 
 \bf lang.     & exp.      & char & word & meta & char & word  & meta  \\ 
\hline
el       & 10     & 96.43& 95.36&\bf 97.01& 0.13 &0.11 & 0.09\\
grc      & 10     & 88.28& 73.52&\bf88.85 & 0.21 &0.29 & 0.22\\
la\_ittb & 10     & 91.45& 87.98&\bf 91.94 & 0.14 &0.30 & 0.05\\
ru       & 10     & 95.98& 93.50&\bf 96.61& 0.06 &0.17 & 0.07\\
tr       & 10     & 93.77& 90.48&\bf 94.67& 0.11 &0.33 & 0.14\\
 \hline
\end{tabular}
\end{center}

\caption{F1 score for the character, word and meta models. The standard deviation of 10 random restarts of each model is show in the last three columns. The differences in means are all statistically significant at {$p<0.001$} (paired t-test). }
\label{tab:ablation-charword}
\end{table}

\paragraph{Concatenation Strategies for the Context-Sensitive Character Encodings}
\begin{table}[t!]
\begin{center}
\small
\setlength{\tabcolsep}{5pt}
\begin{tabular}{|l|c|c|c|c|c|r|l||r|r|r}
\hline 
\bf dev. set.& $\forwardlastnow$ & $\forwardfirstnow$ & $\forwardlastnow$ & $\forwardfirstnow$ & & \\
\bf lang. & $\backwardfirstnow$ & $\backwardlastnow$ & $\backwardlastnow$ & $\backwardfirstnow$ & DQM & $|$xpos$|$\\ 
\hline
el	       &\bf 96.6	& \bf 96.6	&96.2	& 96.1	  & 95.9 &16\\ 
grc	       &\bf 87.3	& 87.1	& 87.1	&86.8     &86.7 &3130\\
la\_ittb   & 91.1	&91.5	&\bf 91.9	&91.3	  &91.0 &811\\
ru         & 95.6    & 95.4	& 95.6	&95.3	  &\bf95.8 &49\\
tr	       & 93.5   & 93.3	&93.2	&92.5     &\bf 93.9  &37\\ \hline
\end{tabular}
\end{center}

\caption{\label{table:char-models} F1 score of char models and their performance on the dev. set for selected languages
with different gather strategies, concatenate to $g_i$ (Equation \ref{eq:gi}).
DQM shows results for our reimplementation of \newcite{DBLP:conf/conll/DozatQM17} (cf.\ \S\ref{sec:word-based}), where we feed in only the characters. The final column shows the number of xpos tags in the training set.}
\vspace{-0.3cm}
\end{table}

The proposed model bases a token encoding on both the forward and the backward character representations of both the first and last character in the token (see Equation \ref{eq:gi}).
Table~\ref{table:char-models} reports, for a few morphological rich languages, the part-of-speech tagging performance of different strategies to gather the characters when creating initial word encodings. The strategies were defined in \S\ref{sec:chars-model}. The Table also contains a column with results for our reimplementation of \newcite{DBLP:conf/conll/DozatQM17}. We removed, for all systems, the word model in order to assess each strategy in isolation. The performance is quite different per language. E.g., for Latin, the outputs of the forward and backward LSTMs of the last character scored highest.

\paragraph{Sensitivity to Hyperparameter Search}

We picked Vietnamese for a more in-depth analysis since it did not perform well and investigated the influence of the sizes of LSTMs for the word and character model on the accuracy of development set. With larger network sizes, the capacity of the network increases, however, on the other hand it is prune to overfitting. We fixed all the hyperparameters except those for the network size of the character model and the word model, and ran a grid search over dimension sizes from 200 to 500 in steps of 50. 
\begin{figure}[t!]
\vspace{-0.7cm}
\includegraphics[scale=0.27]{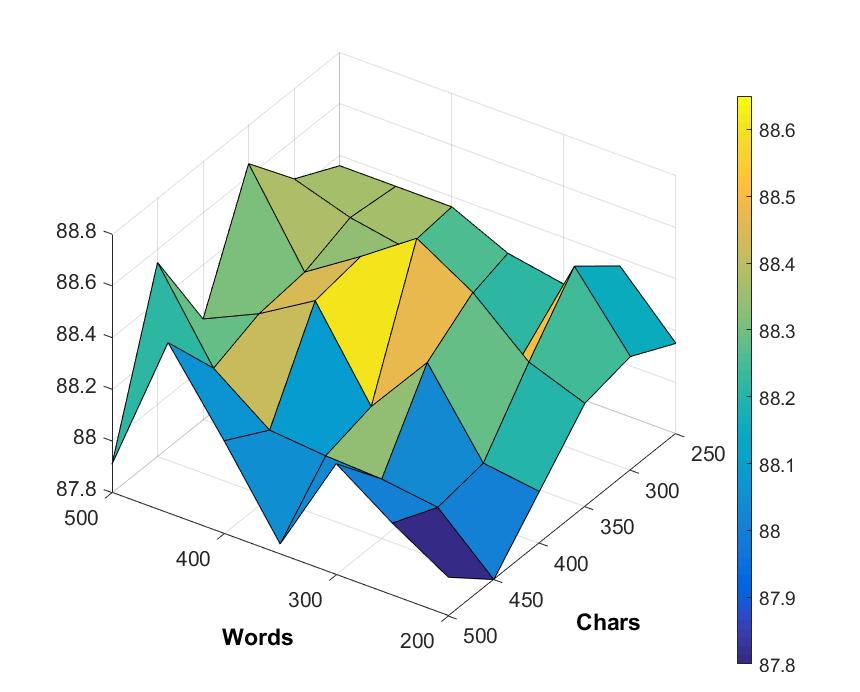}
\caption{3D surface plot for development set accuracy for XPOS (y-axis) depending on LSTM size of the character and word model for the Vietnamese treebank. The snapshot is take after 195 training epochs and we average the values of neighboring epochs.
}
\label{figure:3d-vietnamese}
\vspace{-0.3cm}
\end{figure}
The surface plot in \ref{figure:3d-vietnamese} shows that the grid peaks with more moderate settings around 350 LSTM cells which might lead to a higher accuracy. For all of the network sizes in the grid search, we still observed during training that the accuracy reach a high value and degrades with more iterations for the character and word model. This suggests that future variants of this model might benefit from higher regularization.

\paragraph{Discussion}
\label{sec:discussion}

Generally, the fact that different techniques for creating word encodings from character encodings and different network sizes can lead to different accuracies per language suggests that it should be possible to increase the accuracy of our model on a per language basis via a grid search over all possibilities. 
In fact, there are many variations on the models we presented in this work (e.g., how the character and word models are combined with the meta-BiLSTM). 
Since we are using separate losses, we could also change our training schema. For example, one could use methods like stack-propagation \cite{zhang2016stack} where we burn-in the character and word models and then train the meta-BiLSTM backpropagating throughout the entire network.

\section{Conclusions}
\label{sec:conclusion}

We presented an approach to morphosyntactic tagging that combines context-sensitive initial character and word encodings with a meta-BiLSTM layer to obtain state-of-the art accuracies for a wide variety of languages.

\section*{Acknowledgments}

We would like to thank the anonymous reviewers as well as Terry Koo, Slav Petrov, Vera Axelrod, Kellie Websterk, Jan Botha, Kuzman Ganchev, Zhuoran Yu, Yuan Zhang, Eva Schlinger, Ji Ma, and John Alex for their helpful suggestions, comments and discussions.

\bibliography{acl2018}
\bibliographystyle{acl_natbib}

\end{document}